\documentclass[journal]{IEEEtran}
\usepackage{graphicx}
\usepackage{amssymb,amsmath,bm}
\usepackage{textcomp}
\usepackage{hyperref}
\usepackage{url}
\usepackage[keeplastbox]{flushend}
\usepackage{balance}
\usepackage{subfigure}
\usepackage{xcolor}
\usepackage{pgf}
\usepackage{soul}
\usepackage{threeparttable}
\usepackage{booktabs}
\usepackage{cite}
\usepackage{tikz}
\usetikzlibrary{automata,arrows,positioning,chains,shapes.multipart,fit,petri,shapes,backgrounds,decorations.pathmorphing,calc,shapes.misc, arrows, decorations.markings}
\usepackage{ifthen}

\newcommand{\eg}{e.\,g.,\ }
\newcommand{\ie}{i.\,e.,\ }

\newcommand{\mb}{\mathbf}

\ifCLASSINFOpdf
\else
\fi

\begin{document}
\bstctlcite{IEEEexample:BSTcontrol}

\title{Data Augmentation via Semi-Supervised Conditional Generative Adversarial Networks for Intelligent Health Care}
\title{Data Augmentation via Semi-Supervised Conditional Generative Adversarial Networks}
\title{Automatically Diagnosing Obstructive Sleep Apnea ASSC via Semi-Supervised Conditional Generative Adversarial Networks for Intelligent Health Care}
\title{Snore-GANs: Improving Automatic Snore Sound Classification with Synthesized Data} 

\author{
Zixing~Zhang,~\IEEEmembership{Member,~IEEE,}
Jing~Han,~\IEEEmembership{Student~Member,~IEEE,}
Kun~Qian,~\IEEEmembership{Student~Member,~IEEE,} \\
Christoph Janott,~\IEEEmembership{Student~Member,~IEEE,}
Yanan Guo, 
and~Bj\"orn~Schuller,~\IEEEmembership{Fellow,~IEEE}
\thanks{
This work was supported by the UK's Economic \& Social Research Council through the research Grant No.~HJ-253479 (ACLEW) and the EU's Horizon 2020 / EFPIA Innovative Medicines Initiative through GA No.\,115902 (RADAR-CNS).} 
\thanks{Z.\ Zhang (corresponding author) is with GLAM -- Group on Language, Audio \& Music, Imperial College London, UK (e-mail: zixing.zhang@imperial.ac.uk).}
\thanks{J.\ Han is with ZD.B Chair of Embedded Intelligence for Health Care and Wellbeing, University of Augsburg, Germany (e-mail: jing.han@informatik.uni-augsburg.de).}
\thanks{K. Qian is now with Educational Physiology Laboratory, Graduate School of Education, The University of Tokyo, Japan. He contributed to this work when he was with Machine Intelligence \& Signal Processing Group, MMK, Technical University of Munich, Germany, and also with ZD.\,B Chair of Embedded Intelligence for Health Care and Wellbeing, University of Augsburg, Germany (e-mail: qian@p.u-tokyo.ac.jp). }
\thanks{C.\ Janott is with Institute for Medical Engineering, Technical University of Munich, Germany, and also with audEERING GmbH, Germany (e-mail: cjanott@audeering.com).}
\thanks{Y.\ Guo is with ZD.B Chair of Embedded Intelligence for Health Care and Wellbeing, University of Augsburg, Germany, and also with Lanzhou University, China (e-mail: yanan.guo@informatik.uni-augsburg.de).}
\thanks{B.\ Schuller is with ZD.B Chair of Embedded Intelligence for Health Care and Wellbeing, University of Augsburg, Germany, and also with GLAM -- Group on Language, Audio \& Music, Imperial College London, UK (e-mail: bjoern.schuller@imperial.ac.uk), and also with audEERING GmbH, Germany.}
}

\markboth{}
{Zhang \MakeLowercase{\textit{et al.}}: Snore-GANs: Improving Automatic Snore Sound Classification with Synthesized Data}

\maketitle

\begin{abstract}
One of the frontier issues that severely hamper the development of automatic snore sound classification (ASSC) associates to the lack of sufficient supervised training data. 
To cope with this problem, we propose a novel data augmentation approach based on semi-supervised conditional Generative Adversarial Networks (scGANs), which aims to automatically learn a mapping strategy from a random noise space to original data distribution. The proposed approach has the capability of well synthesizing `realistic' high-dimensional data, while requiring no additional annotation process. To handle the mode collapse problem of GANs, we further introduce an ensemble strategy to enhance the diversity of the generated data. The systematic experiments conducted on a widely used Munich-Passau snore sound corpus demonstrate that the scGANs-based systems can remarkably outperform other classic data augmentation systems, and are also competitive to other recently reported systems for ASSC.
\end{abstract}

\begin{IEEEkeywords}
Snore Sound Classification, Obstructive Sleep Apnea, data augmentation, data synthesis.
\end{IEEEkeywords}

\IEEEpeerreviewmaketitle

\section{Introduction}
\label{sec:introduction}
\noindent
\IEEEPARstart{A}{utomatic} snore sound classification (ASSC) targets at developing an automated and non-invasive method for the classification of Obstructive Sleep Apnea (OSA) based on the snore sound~\cite{Nguyen14-Online,Behar15-SleepAp,Qian16-Classification,Marcel17-Detecting,Sandeep18-Cardiorespiratory}. OSA is characterized by repetitive episodes of decreased (hypopnea) or completely halted (apnea) airflow during sleep, despite the effort to breathe. According to the statistic investigation in~\cite{Punjabi08-epidemiology}, approximately 3$\sim$7\,\% adult men and 2$\sim$5\,\% adult women in the general population around the world suffer from OSA. This leads to a serious deterioration of health conditions, such as daytime sleepiness, excessive fatigue, morning headache, and even high blood pressure and depression mood in a long-term case~\cite{Punjabi08-epidemiology,Karmakar14-Detection,Perez-Macias18-Detection,Yoon18-Slow}. To treat OSA, doctors need to determine the obstructive position of the respiratory tract in the very beginning. 
A standard determination approach often associates with a Drug-Induced Sleep Endoscopy (DISE) procedure, in which a flexible nasopharyngoscope is introduced into the upper airway while the patient is in a state of artificial sleep~\cite{Yoon18-Slow,Janott18-Snoring}. Vibration mechanisms and locations can be observed while video and audio signals are recorded. However, this diagnosis has many disadvantages, such as the exhaustive time-consumption and the high strain of patients~\cite{Janott18-Snoring}. All these disadvantages underline the necessity of ASSC. 

However, the lack of sufficient amounts of labelled data has become one of the major barriers to its progress. 
The rationales behind this problem can be summarized into four points.  
i) \textit{Data privacy}: Due to the sensitivity of health-associated data, patients are often reluctant to publicly share their data. In addition, data privacy regulations restrict the legal usage possibilities of health data~\cite{He18-Privacy}. 
ii) \textit{Time exhaustion when collecting data}. 
For example, to collect less than one thousand labelled samples for the ASSC sub-challenge in the INTERSPEECH 2017 Computational Paralinguistics challenges, about ten years were taken across three hospitals~\cite{Schuller17-INTERSPEECH,Janott18-Snoring}.     
iii) \textit{Imbalanced nature of classes}: In practice, the patients who suffer from a tongue base snoring or an epiglottis snoring are much fewer than the ones from other types of snoring~\cite{Janott18-Snoring}. 
iv) \textit{A High requirement of qualified experts for data annotation}: To label these data, highly experienced experts are demanded to analyze the recorded data and determine the obstruction location based on their prior knowledge. 

The data sparsity problem becomes even worse with the recent rise of high capacity deep neural networks, which are more hungry for data to avoid underestimated parameters and poorly generalized networks~\cite{Zhang17-ADE}.
{\it Data augmentation} is an appealing approach to alleviate the data sparsity problem because it is theoretically able to produce infinite amounts of labelled data at minimum expense. In the context of machine learning, a plethora of data augmentation approaches have been investigated~\cite{Krizhevsky12-ImageNet,Amodei16-Deep,Ko15-Audio}, which generally fall into two groups based on either transformation or synthesis. 
The {\em transformation}-based approaches conduct a certain number of transformation operations on existing samples to generate additional samples while retaining the annotations. These transformation operations include, for example, random cropping, rotation, flips for image samples~\cite{Krizhevsky12-ImageNet}, or the addition of diverse noises for audio samples~\cite{Amodei16-Deep}. Nevertheless, such data augmentation does not improve data distribution which is determined by higher-level features. In contrast, the synthesis-based approaches manage to generate artificial samples given specific labels via a synthesizer. 
The \textit{Synthetic Minority Oversampling Technique (SMOTE)} \cite{Chawla02-SMOTE} is a typical synthesizer-based data augmentation approach, which has been widely used in the domain of machine learning. The underlying idea is the creation of a new set of artificial samples by means of the nearest neighbours belonging to the minority class. 
The problem of the synthesizer-based approaches lies in the realistic gap between the synthetic and real samples, leading the models to learn the wrong information from the synthetic samples. Therefore, improving the synthesizer is considered to be vital to close the gap. 

Over the past few years, a promising generative model, namely {\it Generative Adversarial Networks} (GANs), has attracted extremely widespread research interests in machine learning~\cite{Goodfellow14-GAN,Wang17-GAN,Creswell18-GAN,Han18-Adversarial,Han18-Towards}. 
It consists of two neural networks -- a generator and a discriminator, which contest with each other in a two-player zero-sum game~\cite{Goodfellow14-GAN}. 
Since its inception, GANs have been consistently demonstrated to be powerful in generating impressively realistic images and natural languages~\cite{Goodfellow14-GAN,Radford16-URL,Yu17-SegGAN}. 
In this light, GANs emerge as a potential tool for data augmentation. 
In the literature of machine learning, a handful of related studies have been reported for some applications. 
For instance, for gaze estimation, traditional synthesized images were further decorated by an adversarial network, improving the previous data augmentation models~\cite{Shrivastava17-Learning}. 
For object classification, images were straightforwardly generated by GANs to increase the size of the training set~\cite{Perez17-effectiveness,Antoniou18-Data}, leading to remarkable performance improvement. 
For emotion recognition, several class-specific GANs were used to efficiently transfer data across different domains~\cite{Zhu17-Data}. 

However, no relevant studies have been reported to use GANs to increase the quantity of annotated training data for intelligent health care, especially for the ASSC, to the best of our knowledge.  
Besides, despite the fact that some previous work focuses on synthesizing standalone samples, for example, images, its performance remains unclear in the case of sequential samples, such as audio data, which significantly differs from the standalone samples. The generation of sequential samples, however, heavily relies on the context information~\cite{Yu17-SegGAN}. Albeit a handful of related studies reported in the audio processing domain, they either focus on speech enhancement~\cite{Pascual17-SEGAN,Stoller18-Adversarial} or music creation~\cite{Chen17-Learning}. 

{Motivated by the aforementioned analysis, we made the following contributions in the present article. i) We, for the first time, propose {\em semi-supervised conditional GANs} (scGANs) to generate high-dimensional representations for the ASSC. Compared with classic GANs, the generation process of scGANs is controlled by a condition, and thus there is no need for an additionally exhausting annotation process. } Furthermore, in contrast to conditional GANs~\cite{Zhu17-Data,Sahu17-AAF}, the proposed scGANs require only one model to synthesize different categorical data by the integration of semi-supervised GANs. Besides, when designing the scGANs, we choose the vanilla GANs, rather than other advanced GANs such as Wasserstein GANs~\cite{Arjovsky17-WG}, following the principle of the worst-case scenario and for the sake of easy performance comparison. 
ii) We try to synthesize not only the static acoustic data, but also the sequential acoustic data. For the sequential data, we innovate a recurrent sequence generator with recurrent neural networks, instead of a static data generator.
iii) We introduce an ensemble of GANs to deal with the mode collapse problem. 
iv) We comprehensively investigate three widely used benchmark systems to evaluate the robustness of the proposed methods.  

The remainder of this article is organized as follows. In Section~\ref{sec:dataAug}, we elaborately describe the proposed data augmentation framework with semi-supervised conditional generative adversarial neural networks. {Then, in Section~III, we introduce the database and the experimental setups, followed by the description, analysis, and discussion of the experimental results and findings in Section~\ref{sec:res}. }
Finally, we draw conclusions and suggest future research directions in Section~\ref{sec:conclusion}. 

\section{Methods}
\label{sec:dataAug}
\noindent
{In this section, we first outline the proposed data augmentation framework based on scGANs. Then, we comprehensively describe the principle of GANs and semi-supervised conditional GANs, followed by dynamic alternation and ensemble GANs strategies that are introduced to overcome the training instability and mode collapse problems of scGANs. We finally report the approach to generate acoustic sequences. }

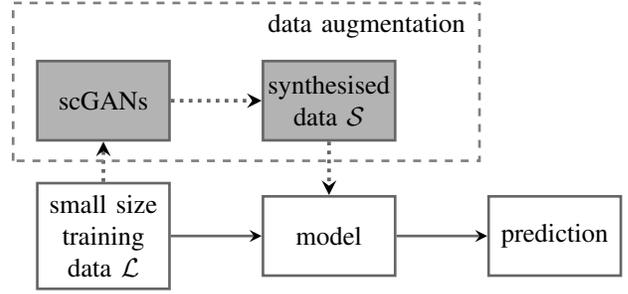
\begin{figure}[t!]
  \centering
%
%
%
%
%

	\def\vsep{1.8cm}
    \def\hsep{3cm} 
   \begin{tikzpicture}[shorten >=0pt,->,>=stealth, draw=black!60, node distance=\hsep, line width=1pt]
   \tikzstyle{object} = [rectangle, draw, fill=white, minimum height=3em, minimum width=5em, inner sep=0pt, align=center];
   \node[object, minimum height=4em] (train_data){small size\\training \\data $\mathcal{L}$};
   \node[object, right of=train_data] (model){model};
   \node[object, right of=model] (prediction){prediction};
   \node[object, above of=train_data, node distance=\vsep, fill=black!30!white] (cgan) {scGANs}; 
   \node[object, above of=model, node distance=\vsep, fill=black!30!white] (syn_data) {synthesised\\data $\mathcal{S}$};
   
   \draw [->] (train_data) to (model);
   \draw [->] (model) to (prediction);
   \draw [->, dotted, line width=1.4pt] (train_data) to (cgan);
   \draw [->, dotted, line width=1.4pt] (cgan) to (syn_data);
   \draw [->, dotted, line width=1.4pt] (syn_data) to (model);
   
   \draw [dashed,gray] (-1.2,1) rectangle (5,3.1);
   \node [node distance=0.5*\vsep] (data) at (3.5,2.8) {data augmentation}; 
   
   \end{tikzpicture}

 \caption{Data augmentation framework based on semi-supervised conditional Generative Adversarial Networks (scGANs) for model training.}
 \label{fig:system}
\end{figure}

\subsection{The Framework of GANs-based Data Augmentation}
\label{subsec:system}
\noindent
The framework of scGAN-based data augmentation is illustrated in Fig.~\ref{fig:system}. In this framework, synthesized data $\mathcal{S}$ is artificially generated through scGANs (see Section~\ref{subsec:scGAN} for more details), and then combined with the original data from a small-sized training set $\mathcal{L}$. The expanded data set, \ie $\mathcal{S}\cup\mathcal{L}$, is further employed to train a model. In this work, we aim to generate high-dimensional representations (features) rather than the raw samples mainly because of the difficulty of learning massive variables in a continuum. 
Under the assumption that the model trained with augmented data shows better performance than the one merely trained with the small-size data set, the simulated data are expected to be able to well reflect the distribution of real data. 

Albeit the availability of some other promising generative models in machine learning, GANs usually empirically outperform them, such as variational autoencoders~\cite{Kingma13-AEV} on the quality of images, and PixelRNN/PixelCNN on the processing speed~\cite{Oord16-Pixel, Oord16-CIG}. 

{
\subsection{Semi-Supervised Conditional GANs}
\label{subsec:scGAN}
\noindent
The vanilla GANs were first introduced in 2014 by Goodfellow~\cite{Goodfellow14-GAN}. They comprise two basic components: a {\it generator} (denoted as $G$) and a {\it discriminator} (denoted as $D$). The $G$ aims to capture the potential distribution of real samples and generates new samples to `cheat' the $D$ as far as possible; whereas the $D$ is often a binary classifier, distinguishing the sources (\ie real samples or generated samples) of the inputs as accurately as possible. 
Therefore, the $G$ and $D$ are normally jointly trained in a two-player zero-sum game, where the total gains of the two players are zero. 
{More details of the vanilla GANs can be found in~\cite{Goodfellow14-GAN}.}

One major problem of the above unconditioned GANs as aforementioned is the lack of label information when generating the data, which constrains its application to data augmentation. {\it Conditional GANs} (cGANs), however, utilize auxiliary information $\mb{c}$, such as the labels or a particular attribute setting, to control the output as desired~\cite{Mirza14-CGA}. 

\begin{figure}[t!]
  \centering
%
%
%
%
%

   \def\layersep{3cm}
   \def\hsep{1cm}
   \begin{tikzpicture}[shorten >=0pt,->,>=stealth, draw=black!60, node distance=\layersep, line width=1pt]
  
      \tikzstyle{object} = [rectangle, draw, fill=white, minimum height=3em, minimum width=5em, inner sep=0pt, align=center];
      \tikzstyle{action} = [circle, draw, text width=0.3cm,font=\small, inner sep=0pt,text centered];
      \tikzstyle{annot} = [rectangle, draw, text width=1em, minimum height=8mm, text centered];
      \tikzstyle{net} = [trapezium, fill=gray!20!white, trapezium angle=75, draw, inner xsep=0pt, outer sep=0pt, minimum height=10mm, text width=6mm,text centered];
      \tikzstyle{dot} = [draw,circle, scale=1];
      \tikzstyle{symbol} = [text centered, text width=1.4cm]; 
      
      \node[annot, text width=1em] (start) at (-0.5,0) {$\mb{z}$};
      \node[symbol, left of=start, node distance=1.1*\hsep] (latent) {latent random vector};  
      \node[net, right of=start, rotate=-90, node distance=1.2*\hsep, fill=blue!30!white] (lstm1) {\mbox{G-net}};
       \node[annot, right of=lstm1, node distance=1.2*\hsep, fill=black!30!white] (pred) {$\hat{\mb{x}}$};
      \node[symbol, above of=pred, node distance=0.9*\hsep] {generated sample}; 
       \node[annot, below of=pred, node distance=0.6*\layersep] (label) {$\mb{x}$};
       \node[symbol, left of=label, node distance=1.2*\hsep] {real~world sample}; 
      \node[net, rotate=90, fill=red!30!white] (lstm2) at (3.5,-0.8) {\rotatebox[origin=c]{180}{D-net}};
      \node[annot, right of=lstm2, node distance=1.5*\hsep,text width=0.4cm, minimum height=2cm] (end) {};
     \node[dot, above of=end, node distance=0.7*\hsep] (dot1) {}; 
     \node[above of=end, rotate=-90, node distance=0.2*\hsep](dots){$\cdots$};
     \node[dot, below of=end, node distance=0.25*\hsep] (dot2) {};
     \node[dot, below of=end, node distance=0.7*\hsep, fill=blue!50!green] (dot3) {};
     \node[right of=dot1, node distance=0.9*\hsep](real_text){class-1};
     \node[right of=dot2, node distance=0.9*\hsep](real_text){class-K};
     \node[right of=dot3, node distance=0.9*\hsep](fake_text){fake};
      \node[symbol, above of=latent, node distance=1.5\hsep] (condition) {condition};  
      \node[annot, above of=start, node distance=1.5\hsep, fill=black!30!green] (condition) {$c$};
      
      \path (start) edge (lstm1) 
      (lstm1) edge (pred)
      (lstm2) edge (end); 
      \draw [->] (pred.east)  to ($(pred.east)+(0.4cm,0cm)$) to ($(pred.east)+(0.4cm,-0.5cm)$) to ($(lstm2.north)+(0cm,0.3cm)$);
       \draw [->] (label.east)  to ($(label.east)+(0.4cm,0cm)$) to ($(label.east)+(0.4cm,0.7cm)$) to ($(lstm2.north)+(0cm,-0.3cm)$);

	   \draw[->, dotted, line width=1.4pt] (condition) -| (lstm1);
       \draw[->, dotted, line width=1.4pt] (condition) -| (lstm2);
      
   \end{tikzpicture}

 \caption{The framework of semi-supervised conditional Generative Adversarial Network (scGANs).}
 \label{fig:scGAN}
\end{figure}
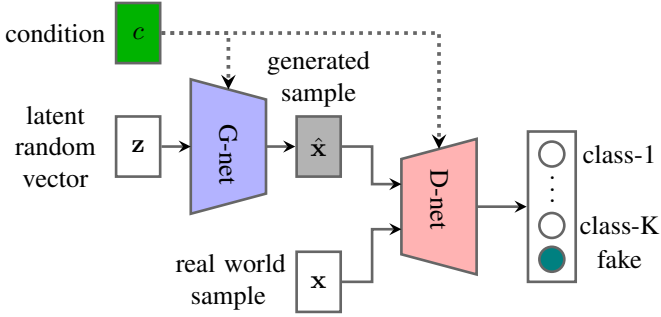

Besides, Odena recently proposed {\it semi-supervised GANs} (sGANs)~\cite{Odena16-Semi}, where the $D$ becomes a combination of a classifier and a discriminator. In detail, the discriminator $D$ classifies the input into $K+1$ classes, where $K$ is the number of classes of a classification task. Real samples are supposed to be classified into the first $K$ classes and the generated samples into the $K+1$-th class (\ie fake). In the framework, however, the generator $G$ aims to generate data that is classified into {\it any} of the first $K$ classes. The benefit of this strategy is two-fold: Firstly, the approach performs well to find the distinguishing boundary, hence creating a data-efficient classifier. Secondly, it empirically performs more efficient for generating higher quality samples than regular GANs~\cite{Odena16-Semi,Salimans16-Improved}. 

Motivated by this work~\cite{Odena16-Semi,Salimans16-Improved}, we propose a novel structure, namely {\it semi-supervised conditional GANs} (scGANs), as structured in Fig.~\ref{fig:scGAN}. They can be considered as extensions of cGANs by forcing the discriminator $D$ to output class labels as well as distinguishing the real data from the fake data. Differing from sGANs~\cite{Odena16-Semi}, the $G$ of the scGANs is conditioned with auxiliary information (\ie label information in this case), and aims to generate data that can be {\it correctly} classified into the first $K$ classes given the condition $c$. 

Mathematically, given real data $\mb{x}$ sampled from the distribution $p_{data}(\mb{x})$, a latent random vector $\mb{z}$ sampled following a simple prior distribution $p_{\mb{z}}(\mb{z})$ (\eg uniform or Gaussian distribution), and the parameters $\mb{\theta}_g$ and $\mb{\theta}_d$ of the $G$ and $D$ networks, respectively, the generator $G$ targets at maximizing the log-likelihood that it assigns to the correct classes:
\begin{equation}
\mathcal{L}_G = \mathbb{E}_{\mb{z}\sim p_{\text{c}}(\mb{z})}[\log P(y=k|\hat{\mb{x}})], 
\end{equation}
whilst the discriminator $D$ aims to maximize the following log-likelihood: 
\begin{equation}
\begin{aligned}
\mathcal{L}_D = & \mathbb{E}_{\mb{x}\sim p_{\text{data}}(\mb{x})}[\log P(y=k|\mb{x})] +  \\
& \mathbb{E}_{\mb{z}\sim p_{\text{z}}(\mb{z})}[\log P(y=fake|\hat{\mb{x}})], 
\end{aligned}
\end{equation}
where $k$ is among the first $K$ classes, $\hat{\mb{x}}=G_{\mb{\theta}_g}(\mb{z}|\mb{c})$, and $p_{c}(\mb{z})$ indicates the latent random distribution relating to the conditional information $\mb{c}$. 
By taking the class distribution into the objective function, an overall improvement in the quality of the generated samples is expected. 

It has to be noticed that the proposed scGANs differ from the ones in~\cite{Sricharan17-Semi}, which are structured with two discriminators, used for unsupervised (with unlabelled real data) and supervised (with class-specific real data) true/false classification, respectively. The proposed scGANs, however, can be further extended with two discriminators in case of exploiting unlabelled data in future efforts, which is beyond the research scope of this article.

\subsection{Dynamic Alternation and Ensemble of Semi-Supervised Conditional GANs}
\label{subsec:ensemblescGAN}
Generally, the training of $G$ and $D$ is conducted in an iterative manner, \ie the  corresponding neural weights $\mb{\theta}_d, \mb{\theta}_g$ are updated in turns~\cite{Goodfellow14-GAN}. Once the training is completed, the generator $G$ is able to generate more realistic samples, while the discriminator $D$ can distinguish authentic data from fake data. 
The adversarial training process, however, suffers from two major issues: {\it training instability} and {\it mode collapse}~\cite{Wang17-GAN,Creswell18-GAN}. 

When training the adversarial networks, ensuring the balance and synchronization between the $G$ and $D$ plays an important role in obtaining reliable results~\cite{Goodfellow14-GAN}. That is, the optimization goal of adversarial training lies in finding a saddle point of, rather than a local minimum between $G$ and $D$. The inherent difficulty in controlling the synchronization of the two adversarial networks increases the risk of {\it instability} in the training process. 

In this light, we introduce a simple and efficient way called {\it dynamic alternation}. That is, we dynamically alternate the training epochs between the generator $G$ and the discriminator $D$, in contrast to the conventional approaches which often fix the training epochs for both (fixed alternation). It is hoped that this approach is able to keep the learning pace synchronously updated between $G$ and $D$, so as to avoid the training instability.  

Mathematically, we respectively define a loss threshold function for $G$ and $D$ with 
\begin{equation}\label{eq:dynThres}
    \mathcal{L}_{TG/D} = \text{max}(\Lambda^{i} + b, c), 
\end{equation}
where $\Lambda$, $b$, and $c$ are the hyper-parameters which control the threshold together at the $i$-th training iteration. To guarantee $\mathcal{L}$ being a monotonically decreasing function, $\Lambda$ is normally less than 1. 
In this article, these hyper-parameters are determined by empirical experience. 
Once the training loss from $G$ is below a pre-defined loss $\mathcal{L}_{TG}$, the training process is altered to $D$. Similarly, once the training loss from $D$ is below another pre-defined loss $\mathcal{L}_{TD}$, the training process is altered to $G$. Such an alternation keeps repeating until a training convergence of $G$ and $D$. In doing this, we force the performance improvement of $G$ and $D$ at a similar pace. 

\begin{figure}[t!]
  \centering
%
%
%
%
%

	\def\vsep{1.5cm}
    \def\hsep{3cm} 
   \begin{tikzpicture}[>=latex,node distance=\vsep]
   \tikzstyle{data}=[cylinder, shape border rotate=90, aspect=0.2, draw, minimum height=10mm, minimum width=20mm, inner sep=0pt, align=center] {};
   \tikzstyle{object} = [rectangle, draw, fill=white, minimum height=3em, minimum width=5em, inner sep=0pt, align=center, node distance=\hsep];
   \tikzstyle{myarrow}=[line width=1mm,draw=black!20!white,-triangle 45,postaction={draw, line width=2mm, shorten >=3mm, -}]

   \foreach \i in {1,5}{
     \ifthenelse{\i = 1}
     {\node[object](cgan-\i) at (\i, 0) {scGANs-1}}
     {\node[object](cgan-\i) at (\i, 0) {scGANs-n}};
     \node[data, below of=cgan-\i](data-\i) at (\i, 0) {synthesised\\ data};
   }
   \node[right of=cgan-1, node distance=2cm](dot-1) {$\cdots$};
   \node[below of=dot-1](dot-2) {$\cdots$};
   \node[data, below of=dot-2, node distance=2.2cm, minimum width=20mm, fill=blue!40!white](data) {pool of\\synthesised\\data};
   
   \node[circle, draw, inner sep=2pt, below of=data, node distance=2cm](add) {\large{$+$}}; 
   
   \node[object, left of=add, minimum width=30mm] (train_data) {original small size\\training data}; 
   \node[object, right of=add, minimum width=25mm, fill=black!20!white] (new_data) {augmented \\training data}; 
   
   \draw [->] (cgan-1) to (data-1);
   \draw [->] (cgan-5) to (data-5);
   \draw [myarrow] ($(data-1.south)+(0,-0.3)$) -- (data) node[midway, sloped, below, node distance=2cm] {added to};
   \draw [myarrow] ($(data-5.south)+(0,-0.3)$) -- (data) node[midway, sloped, below] {added to};
   
   \draw [->, line width=2pt, color=gray] (data) -- (add) node[midway, right] {randomly select};
   \draw [->, line width=2pt, color=gray] (train_data) to (add); 
   \draw [->, line width=2pt, color=gray] (add) to (new_data); 


   \end{tikzpicture}

 \caption{Data augmentation by using an ensemble of semi-supervised conditional Generative Neural Networks (scGANs). n: the number of scGANS. }
 \label{fig:ECGAN}
 \vspace{-.3cm}
\end{figure}
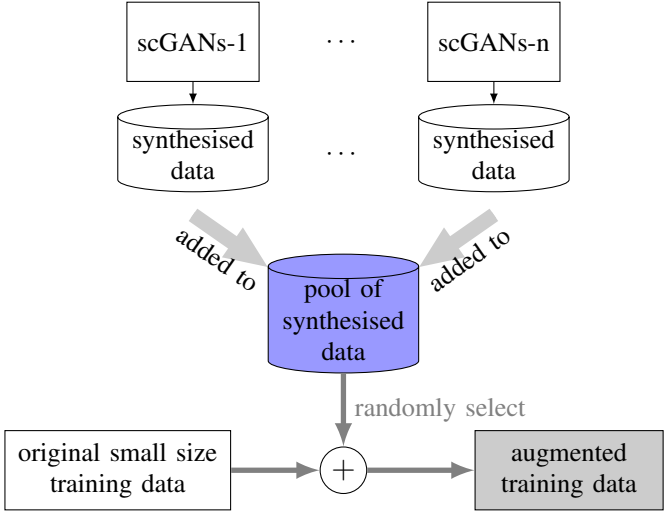

Apart from the training instability, another issue is the {\it mode collapse}, which indicates that the generated samples have integrated into a small subset of similar samples (partial collapse), or even a single sample (complete collapse). In this case, the $G$ exhibits  very limited diversity amongst generated samples, thus reducing the usefulness of GANs. 

To address this problem, some approaches are continually emerging. For example, the cost function of the generator can be modified to factor the diversity of generated batches~\cite{Salimans16-Improved}. Moreover, the unroll-GANs allow the generator to `unroll' updates of the discriminator in a manner which is fully differentiable~\cite{Metz16-UGA}.

More recently, the work shown in~\cite{Wang16-Ensembles}, especially its advanced version~\cite{Tolstikhin17-Adagan}, demonstrated that an ensemble of several networks with different network structures or initializations can improve the system performance significantly, in comparison with the aforementioned unroll-GANs~\cite{Wang16-Ensembles, Tolstikhin17-Adagan}. 
To this end, we implement a standard ensemble approach~\cite{Wang16-Ensembles} in our experiments for the sake of easy comparison, hoping to obtain a better estimation of the real data distribution $p_{data}$. The framework of the ensemble of scGANs for data augmentation is depicted in Fig.~\ref{fig:ECGAN}.   
Instead of training a single scGANs pair, we train a set of scGANs. These scGANs are with different network structures (\ie a different number of hidden nodes per layer in our experiments) in order to maximally explore their differences, and trained independently. When conducting data augmentation, we aggregate the data from all scGANs, and randomly select data from the pool which are further merged into the original training set. By doing this, it is expected to expand the diversity of the augmented data that come from separate scGANs. 
}

\subsection{Sequence Generation} 
\label{subsec:seqGen}
\noindent
The snore data are normally structured in a sequence. 
However, most available GANs were particularly designed to generate standalone samples (\eg images). 
In this section, we introduce a novel approach to generate sequential samples by means of the GANs equipped with Recurrent Neural Networks (RNNs) with Gated Recurrent Units (GRUs), since the GRU-RNNs have been widely known to be efficient in capturing long-range context information~\cite{Chung14-Empirical, Zhang17-Learning, Zhang18-Evolving}. 

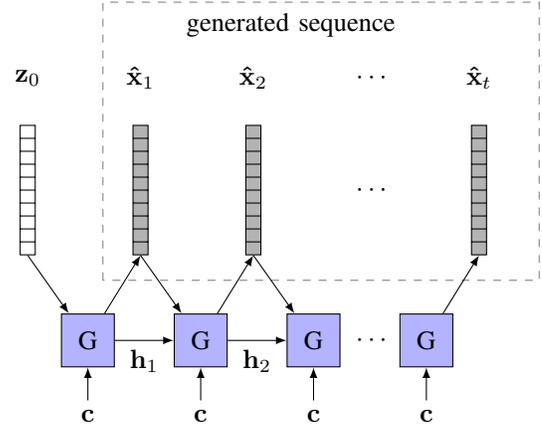
\begin{figure}[t!]
  \centering
%
%
%
%
%

   \begin{tikzpicture}[>=latex,node distance=0pt]
	\tikzstyle{vector}=[rectangle split,rectangle split parts=10,draw,text centered, minimum height=2cm, minimum width=0.2cm, inner ysep=0pt,  inner xsep=0pt, fill=black!30!white]
    \tikzstyle{generator} = [rectangle, draw, fill=blue!30!white, minimum height=2em, minimum width=2em, inner sep=0pt, align=center];
    
    \foreach \x [count=\n] in {0,1,2,4}{
    	\ifthenelse{\x = 0}
        {\node[vector, fill=white](v-\x) at (1.5*\x,0){}}
        {\node[vector](v-\x) at (1.5*\x,0){}};
        \ifthenelse{\x = 0}{\node[](x-\x) at (1.5*\x,1.5){$\mb{z}_0$}}{};
        \ifthenelse{\x = 1}{\node[](x-\x) at (1.5*\x,1.5){$\mb{\hat{x}}_\x$}}{};
        \ifthenelse{\x = 2}{\node[](x-\x) at (1.5*\x,1.5){$\mb{\hat{x}}_\x$}}{};
        \ifthenelse{\x = 4}{\node[](x-\x) at (1.5*\x,1.5){$\mb{\hat{x}}_t$}}{};
        \ifthenelse{\x = 4}
        {\node[generator](g-\x) at (1.5*\x-0.7,-2){G}}
        {\node[generator](g-\x) at (1.5*\x+0.8,-2){G}};
	}
    \foreach \x [count=\n] in {-2,0,1.5}{
    \node[](d-\x) at (4.6, 1*\x){$\cdots$}; 
    }
    
    \foreach \x [count=\n] in {0,1,2,4}{
    \node[below of=g-\x, node distance=1cm](c-\x){$\mb{c}$}; 
    \draw [->] (c-\x) to (g-\x);
    }
    
   \draw [->] (g-0) -- (g-1) node[midway, below] {$\mb{h}_1$};
   \draw [->] (g-1) -- (g-2) node[midway, below] {$\mb{h}_2$};
    
   \draw [dashed,gray] (1,-1.2) rectangle (6.8,2.5);
   \node [] (seq) at (3.5,2.2) {generated sequence}; 
    
   \draw [->] (v-0.south) to (g-0);
   \draw [->] (v-1.south) to (g-1);
   \draw [->] (v-2.south) to (g-2);
   \draw [<-] (v-1.south) to (g-0);
   \draw [<-] (v-2.south) to (g-1);
   \draw [<-] (v-4.south) to (g-4);
   
   \end{tikzpicture}

  \caption{Sequence generation through a recurrent generator ($\mb{h}_t$: hidden states at time $t$, $\mb{c}$: conditional vector).}
 \label{fig:seqGen}
 \vspace{-.3cm}
\end{figure}

To be more specific, given a sequence $\mb{x}_{1:T}=\{\mb{x}_1, \mb{x}_2,\ldots, \mb{x}_T\}$, which comprises of $T$-length consecutive high-dimensional vectors $\mb{x}$, the goal of the recurrent GANs is to learn 
\begin{equation}
f(\mb{x}_t|\mb{x}_1,\ldots,\mb{x}_{t-1}; \mb{c}), 
\end{equation}
while $\mb{x}_1 = f(\mb{z};\mb{c})$. 
In doing this, they are able to generate a complete sequence $\hat{\mb{x}}_{1:T}$ by feeding a latent random noise $\mb{z}$, \ie $\hat{\mb{x}}_{1:T}=G(\mb{z})$. Intuitively, we illustrate the sequence generation process in Fig.~\ref{fig:seqGen}. 

The generation process is indeed inspired by the Seq2Seq modeling~\cite{Sutskever14-Sequence}, where the decoder component takes the last output at time $t-1$ as its input at time $t$, and takes the previous hidden state at time $t-1$ as its initial state at time $t$. 
Differently, the conditional vector $\mb{c}$ is consistently used to guide $G$ to produce a designed sequence. 

As to the discriminator $D$, we further utilize another GRU-RNN to distinguish the generated sequences from the real ones.

{\section{Training and Validation}}
\label{sec:exper}
\noindent
To evaluate the performance of the proposed data augmentation approaches for ASSC, we selected the Munich-Passau Snore Sound Corpus (MPSSC). The corpus has been widely used in the intelligent health care research community~\cite{Qian16-Classification,Janott18-Snoring}, and has been employed as an official database for an ASSC sub-challenge in the INTERSPEECH 2017 Computational Paralinguistics challenges~\cite{Schuller17-INTERSPEECH}. 

\subsection{The Munich-Passau Snore Sound Corpus} 
\label{subsec:dataset}
\noindent

The MPSSC was introduced to classify the vibration location within the upper airways when snoring~\cite{Qian16-Wavelet,Qian16-Classification,Janott18-Snoring}. Since the data analysis and annotation process did not involve any studies carried out by the authors on humans or animals, an ethical approval was not required. Starting material for the database were existing recordings of DISE examinations from three medical centres in Germany (\ie Klinikum rechts der Isar, Technical University Munich; Alfried Krupp Hospital Essen, and University Hospital Halle/Saale) which were taken during clinical routine examinations between 2006 and 2015, using different recording devices among the medical centres. In a DISE examination, the patient is slightly sedated and put into a condition that resembles an artificial sleep state. By means of a flexible nasopharyngoscope, the upper airways are observed by an experienced ENT physician, identifying the locations of tissue vibration or airway narrowing while the patient snores or undergoes obstructive events. Both the audio signals from the microphone and the video signals from the nasopharyngoscope were recorded synchronously. For the database, in excess of 30 hours of DISE recordings were analyzed.  

\begin{table}[!t]
\caption{data distribution of the Munich-Passau Snore Sound Corpus (MPSSC). V: Velum; O: Oropharynx; T: Tongue; E: Epiglottis}
\centering
\begin{threeparttable}
\begin{tabular}{lrrrr}
\toprule
{\#} & {train} & {devel}  & {test} & {$\sum$} \\ 
\midrule
{V} & {161} & {168} & {155} & {484} \\ 
{O} & {75} & {76} & {65} & {216} \\ 
{T} & {15} & {8} & {16} & {39} \\ 
{E} & {32} & {30} & {27} & {89} \\ 
\midrule
{$\sum$} & {283} & {282} & {263} & {828} \\ 
\bottomrule
\end{tabular}
\end{threeparttable}
\label{tab:snore}
\vspace{-.2cm}
\end{table}

{For our experiments, the audio signal was extracted from the mp4 recordings and stored in a
wav-format (16 bit, 44.1\,kHz). To detect the snore sound events from the audio files, an automated algorithm was employed. In details, 
we averaged the absolute value of the signal amplitude in 10\,ms segments with no overlap and determined the background noise level by means of a 1024-step histogram averaging 10\,s segments~\cite{Janott18-Snoring}. Only the segments, which exceed two times the predefined background noise level for a minimum duration of 300\,ms, were annotated. After that, we added 100\,ms of signals before and after the actual onset and end of the event, which was then extracted from the original audio file, normalized, and saved as separate wav files (16 bit, 16\,kHz)~\cite{Janott18-Snoring}. Finally, an experienced human listener (the fourth author) listened to all selected events and classified them manually as either pure snoring (snore) or other sounds (non-snore, or the snore severely disturbed by non-static background noise)~\cite{Janott18-Snoring}. For more details of this pre-processing step, please refer to~\cite{Janott18-Snoring}.  
}

The selected snore events were then classified by medical ENT (ear, nose, and throat) experts based on the findings from video recordings. Only events with a clearly identifiable (\ie single site of vibration and without obstructive disposition) were included in the database, resulting in 828 snore events in total. 
Based on the VOTE scheme that is widely used to distinguish four structures involved in airway narrowing and obstruction~\cite{Kezirian11-Drug}, we defined four classes: V (\ie Velum, including soft palate, uvula, lateral velopharyngeal walls), O (\ie Oropharyngeal lateral walls, including palatine tonsils), T (\ie Tongue, including tongue base and airway posterior to the tongue base), and E (\ie Epiglottis). 

The annotated audio samples were then separated into subject-independent training, development, and test partitions. Table~\ref{tab:snore} displays the data distribution by partitions and classes. The database is strongly imbalanced with a comparatively low number of T and E samples. This is consistent with earlier medical research, which has found that respiratory disturbances occur more frequently at velopharyngeal and oropharyngeal level, compared to the hypopharyngeal level \cite{Hessel02-Diagnostic}. For more specific information about the database, the readers are referred to~\cite{Janott18-Snoring}.

\subsection{Representations} 
\label{subsec:representations}
\noindent
To keep in line with the ASSC benchmark of the 2017 INTERSPEECH Computational Paralinguistics challenges~\cite{Schuller17-INTERSPEECH}, we chose three different kinds of acoustic feature sets at either the frame level (\ie low-level descriptor) or the segment level (\ie functional or Bag-of-Audio-Words). 

\subsubsection{Low-Level Descriptors} 
\label{subsec:llds}
\noindent
We used the ComParE16 high-dimensional acoustic feature set employed in~\cite{Schuller16-INTERSPEECH}, which contains 65 frame-wise Low-Level Descriptors (LLDs, \eg energies, Mel-frequency cepstral coefficients, zero-cross rate, jitter, shimmer, probability of voicing) as well as their first derivations, leading to 130 LLDs. 
These LLDs are determined according to a set of brute-force empirical evaluations on computational paralinguistics~\cite{Schuller16-INTERSPEECH,Schuller17-INTERSPEECH}. More detailed information about the ComParE16 LLDs can be found in Table~\ref{tab:compare_llds}. 

\begin{table}[!t]
\renewcommand{\arraystretch}{0.8}
\caption{The \textsc{ComParE} acoustic feature set includes 65 low-level descriptors (LLDs) of different types, as well as their first derivations (delta), resulting in 130 LLDs.}
\vspace{-3mm}
\centering
\begin{threeparttable}
\begin{tabular}{@{}l|l@{}}
\toprule
\textbf{4 energy-related LLDs}  & \textbf{Group}\\ 
\midrule
RMS energy, zero-crossing rate &  Prosodic \\
sum of auditory spectrum (loudness)  & Prosodic \\
sum of RASTA-filtered auditory spectrum  & Prosodic \\ 
\midrule
\textbf{55 spectral LLDs} & \textbf{Group}\\ 
\midrule
MFCC 1--14 & Cepstral\\
psychoacoustic sharpness, harmonicity & Spectral\\
RASTA-filt.\ aud.\ spect.\ bds.\ 1--26 (0--8\,kHz) & Spectral\\
spectral energy 250--650\,Hz, 1\,k--4\,kHz & Spectral\\
spectral flux, centroid, entropy, slope & Spectral\\
spectral roll-off point 0.25, 0.5, 0.75, 0.9 & Spectral\\
spectral variance, skewness, kurtosis & Spectral \\ 
\midrule
\textbf{6 frequency-related LLDs} & \textbf{Group}\\ 
\midrule
$f_0$ (SHS and Viterbi smoothing) & Prosodic\\
probability of voicing & Voice quality \\
log.\ HNR, jitter (local and $\delta$), shimmer (local) & Voice quality\\
\bottomrule
\end{tabular}
\end{threeparttable}
\vspace{-.4cm}
\label{tab:compare_llds}
\end{table}

\subsubsection{Functional-based Features}
\label{sec:functionals}
Intuitively, the functional-based approach projects the temporal LLD contours onto a set of feature vectors with descriptive statistic functionals (see~\cite{Eyben10-openSMLE} for more details). Mathematically, this can be written as follows: 
\begin{equation}
 \mb{f} = f([\mb{x}_{i}], i = 1, \ldots, T),
\end{equation}
where  $\mb{f}$ denotes the segment-level feature vector; $[\mb{x}_i]$ indicates the sequential frame-wise LLDs; $T$ is the total frames of a given vocalization; and $f$ denotes the {\it functionals} (\ie statistic information) that are applied per LLD contour.
Specifically, the functionals can include: extremes (minimum, maximum, ranges, etc.), mean (arithmetic, quadratic, geometric), moments (variance, skewness, kurtosis, etc.), percentiles (quantiles, ranges, etc.), peaks (number, distances, etc.), temporal variables (durations, positions, etc.), and regression (coefficients, error).
For our experiments, the functional-based feature set contains 6\,373 dimensional feature vectors~\cite{Schuller16-INTERSPEECH}.

\subsubsection{Bag-of-Audio-Words} 
Bag-of-Audio-Words (BoAW) is another type of segment-level acoustic representation. 
Extracting BoAW involves three steps: i) {codebook generation}; ii) {vector quantization}; and iii) {histogram construction}. 
Differing from bag-of-words for linguistic analysis, the total number of audio-words (frame-wise LLDs) is indeed numerous with an equal occurrence frequency of one. 
To reduce the codebook size ($S$), a $k$-means clustering or a random sampling is conducted to determine the codewords ($W$) of the codebook ($C$)~\cite{Schmitt17-openxbow}. 
After that, a multi-assignment quantization technique is executed to map each audio-word to the first $n$ closest codewords, measured by Euclidean distance. 
Finally, a histogram is constructed by calculating the counts of occurrence of each codeword in all acoustic frames over one vocalization segment. Mathematically, the histogram representation $\mb{b}$ for a given vocalization $v$ with $T_v$ frames is 
\begin{equation}
 \mb{b} = [\sum_{i=1}^{T_v}{\phi_{i,m}}], {m=1, \ldots, S}, 
\end{equation}
where $\phi_{i,m}$ equals to 1 if the $i$-th frames is assigned to the $m$-th codeword, otherwise, to 0. 
To minimize the effects relating to the length disparities of different vocalizations, a normalization process is further undertaken over $\mb{b}$, to sum up all elements of $\mb{b}$ to one. More details about the BoAW generation can be found in~\cite{Schmitt17-openxbow}. 

\subsection{Implementation Setups} 
\label{subsec:setups}
\noindent
As to the acoustic features, we utilized the open-source toolkit {\it openSMILE}~\cite{Eyben10-openSMLE} to extract the LLDs and functional-based features, and the toolkit {\it openXBOW}~\cite{Schmitt17-openxbow} to distill the BoAW representations. 

When simulating the representations, we deployed GRU-RNN-based scGANs. {The generator and discriminator used the same network structure, with two hidden layers and $N$ nodes per hidden layer, where $N$ was set to be 60. As to the discriminator, we appended an additional dense layer and a softmax activation function for pattern classification. As to the GRUs, we employed the standard version with sigmoid and tangent activation functions~\cite{Chung14-Empirical}.}
{To train the networks, we employed the Adam optimization algorithm with an optimized learning rate of 0.001 for the generator and 0.01 for the discriminator.} The batch size was set to be 64 to facilitate the training process. To improve the generalization of the neural networks, we further applied an L2 regularization term to the loss function with a regulation value of 10E-4. {Note that all these hyper-parameters were optimized on the development set with the baseline system -- the one without GAN-based data augmentation and taking LLD acoustic features as inputs. Thus, it avoids exhausting computation caused by the grid-searching in numerous experimental scenarios. Besides, we set the initial state of GRU to be zero, and the initial weights of neural networks to be random values with a standard deviation of 0.1.} 
{To find the saddle point between the generator and discriminator when training the GANs, we employed the dynamic alternation strategy as described in Section~\ref{subsec:ensemblescGAN} to alternatively train the generator and discriminator. 
Specifically, with respect to Eq.~(\ref{eq:dynThres}), we set $\Lambda$, $b$, and $c$ to be 0.95, 0, and 0.7 for the discriminator, and 0.95, 1.0, and 1.0 for the generator. These hyper-parameters were set according to empirical experience. In further, we could use more advanced approaches to search these values, for example, reinforcement learning~\cite{Zhang18-Evolving}.}

For the sequence generation, we partitioned the original sequence with variable length into multiple continuous segments with a fixed window size of 400\,ms and a step size of 100\,ms. This partially reduces the complexity of sequence generation. 

Due to the distinct characteristics of the three investigated features (cf.~Section~\ref{subsec:representations}), we considered one static model, \ie Support Vector Machines (SVMs), which aims to learn the segmental-level features (\ie based on functionals or BoAWs), and one dynamic model, \ie GRU-RNNs, which attempts to learn the sequential frame-level LLDs. 
In our experiments, three learning systems have been implemented, referring to i) functional-based features with SVMs ({\it functionals + SVMs}), ii) BoAW-based features with SVMs ({\it BoAWs + SVMs}), and iii) sequential LLDs with GRU-RNNs ({\it LLDs + GRU-RNNs}), respectively. 
Particularly, the selection of SVMs rather than other typical classifiers mainly relates to two reasons: i) SVMs have been officially employed in the 2017 INTERSPEECH ASSC sub-challenge~\cite{Schuller17-INTERSPEECH}, as well as other related studies~\cite{Wiens17-Machine,Jiang17-Artificial,Beam18-Big}; ii) our previous experimental results have shown that SVMs generally perform more stable and better than other typical classifiers (\eg random forest) on the MPSSC database. 
For example, we obtained the UARs on the development and test sets of 34.9\,\% and 35.2\,\% by using ‘functional+RF’ system, and 38.9\,\% and 55.2\,\% by using ‘BoAW + RF’ system. These results are generally inferior to the obtained results by SVMs (cf.~Table~\ref{tab:da}). 

As to the SVMs, the complexity  was determined on the development set through the baseline experiments without data augmentation by searching values among $[0.00001, 0.00005, 0.0001 \ldots, 0.5, 1, 5]$. 
Empirically, the complexity was optimized to be $10E-4$ and $10E-3$ in the cases of functional-based features and BoAW representations, respectively. For the GRU-RNNs model, we employed the same network structure as for the discriminator in the scGANs, but with only four output nodes (the discriminator has five output nodes due to the `fake' prediction). When training the GRU-RNNs, a many-to-one strategy was used, \ie after feeding a sequence of LLD vectors, only the last states from the hidden layers were considered for final classification. Again, the Adam optimizer was implemented with the same learning rate and L2 regularization value with the parameters of the discriminator. 

To evaluate the performance of the investigated systems, we 
{kept in line with the evaluation metric, \ie {\em Unweighted Average Recall} (UAR), which was officially employed in the 2017 INTERSPEECH ASSC sub-challenge, for the sake of performance comparison. }  
The UAR is calculated by the sum of recalls per class divided by the number of classes, and thus can reflect a meaningful overall accuracy despite class imbalances, such as the one we are facing. 

{Due to the imbalanced class distribution of the MPSSC database, we oversampled the data from the minority classes by means of replication, forcing an even distribution. Different from the proposed data augmentation approach that aims to synthesize completely new data, this strategy increases the weights of the losses from the minority samples. The advantage relates to the fact that it increases the contributions of the original minority samples which hold grounded class information when modeling ASSC.}
Notably, when using GRU-RNNs to classify each recording, a majority voting strategy was applied to a set of related segments to come up with a final prediction, since each recording was split into several sub-segments as aforementioned. 


{\section{Results and Discussion}
\label{sec:res}
\noindent
In this section, we conducted comprehensive evaluations of the systems with proposed scGAN-based data augmentation (snore-GANs) on the selected MPSSC database.
}

{
\subsection{Dynamic Alternation Evaluation}
\label{subsec:res_dlt}
\noindent
Before the systematic performance evaluation, we firstly investigated the efficiency of the introduced dynamic alternation training strategy for the scGANs. In Fig.~\ref{fig:res_dynamic}, we plotted the obtained losses at each learning iteration of both the generator $G$ and the discriminator $D$. From the figure, we can see that when using the conventional fixed-alternation training strategy, the obtained loss curves from both $G$ and $D$ severely vibrate along with the learning iterations, which clearly shows the training instability when the number of epochs per iteration is fixed for $G$ and $D$ (see Fig.~\ref{fig:res_dynamic} (a)). In contrast, they are shown to be much smoother when using the dynamic alternation training strategy (see Fig.~\ref{fig:res_dynamic} (b)). This suggests that the dynamic alternation training strategy is capable of improving the training stability of GANs and thus facilitates the convergence of the training process, by forcing the learning process of both networks to be in a similar pace. Nevertheless, it is worth noting that to determine the hyper-parameters of the dynamic loss threshold requires empirical experience. 

\begin{figure}[t!]
    \centering
    \subfigure[fixed alternation]{\includegraphics[width=1.7in,height=1.5in]{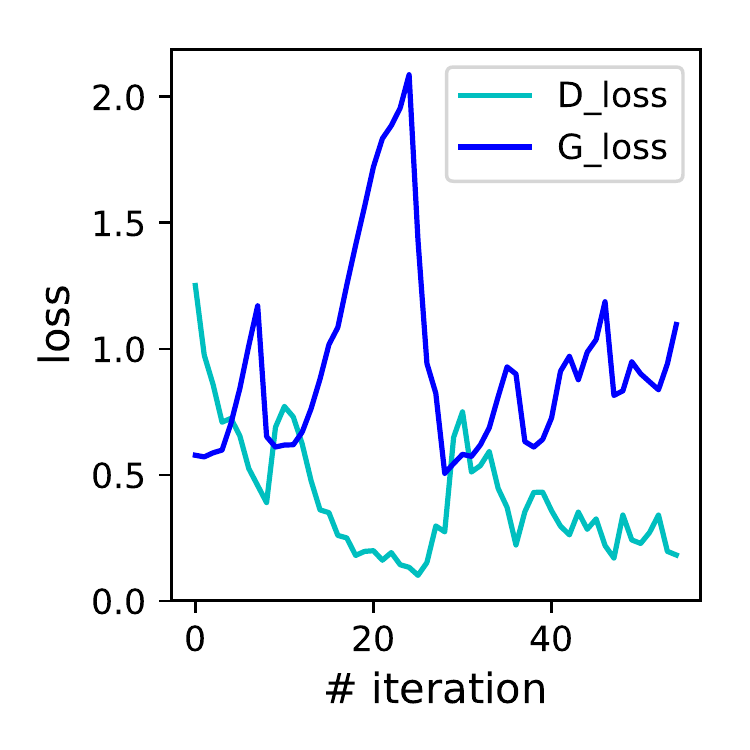}}
    \subfigure[dynamic alternation]{\includegraphics[width=1.7in,height=1.5in]{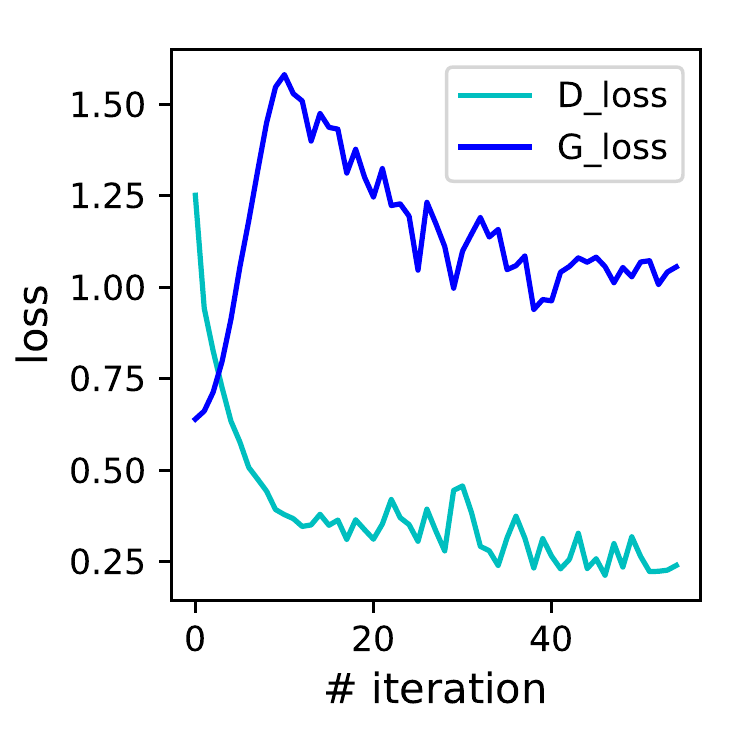}}
    \caption{{The variation of losses while training the scGAN at every iteration using the fixed alternation strategy (a) or the dynamic alternation strategy (b).}}
    \label{fig:res_dynamic}
    \vspace{-.2cm}
\end{figure}
}

\subsection{Results of scGANs and Discussion}
\noindent
When generating the data, we filtered out such samples that are not correctly recognized by the discriminator. In doing this, it potentially removes the noisy samples possibly falling beyond the scope of the original data distribution, and thus alleviates their adverse effect in learning. Moreover, when adding the generated data to the original training set, we randomly selected the generated samples evenly distributed over categories, in order to handle the imbalanced data distribution problem. 

\begin{figure*}[t!]
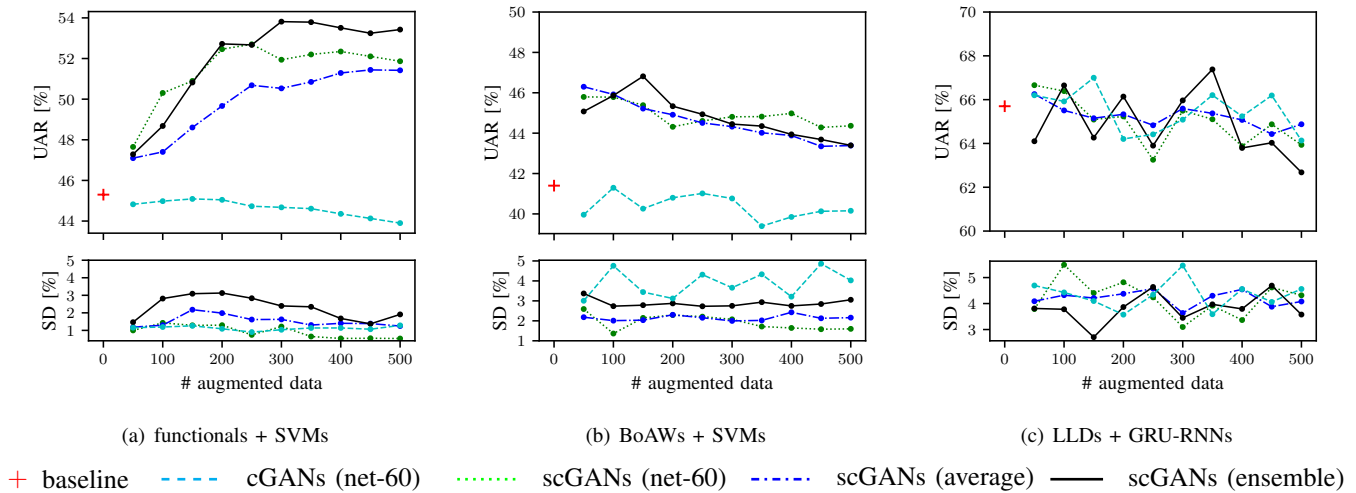

   \centering
    \subfigure[{functionals + SVMs}]{
      \resizebox{0.31\textwidth}{2.2in}{\input{da_ComParE16_func_utt.pgf}}
    }
    \subfigure[{BoAWs + SVMs}]{
      \resizebox{0.31\textwidth}{2.2in}{\input{da_ComParE16_BoAW_utt.pgf}}
    }
    \subfigure[{LLDs + GRU-RNNs}]{
      \resizebox{0.31\textwidth}{2.2in}{\input{da_ComParE16_lld_win.pgf}}
    }
%
%
%
%
%

   \def\hsep{4cm}
   \begin{tikzpicture}[shorten >=0pt, draw=black!60, node distance=\hsep, line width=1pt]
  
      \tikzstyle{symbol} = [text centered, text width=1cm]; 
      \node[color=red] (add) {$+$};
      \node[] (label1) at (0.9,0) {baseline}; 
      \node[right of=label1, node distance=3.3cm] (labelc) {cGANs (net-60)};
      \node[right of=labelc] (label2) {scGANs (net-60)};
      \node[right of=label2] (label3) {scGANs (average)};
      \node[right of=label3] (label4) {scGANs (ensemble)};

      \draw [color=cyan, dashed] ($(labelc.west) + (-1,0)$)  --  ($(labelc.west) + (-0.2,0)$); 
      \draw [color=green, dotted] ($(label2.west) + (-1,0)$)  --  ($(label2.west) + (-0.2,0)$); 
      \draw [color=blue, dashdotted] ($(label3.west) + (-1,0)$)  --  ($(label3.west) + (-0.2,0)$); 
      \draw [color=black] ($(label4.west) + (-0.9,0)$)  --  ($(label4.west) + (-0.2,0)$); 
      
   \end{tikzpicture}

   \caption{{Performance (UAR as well as corresponding Standard Deviation [SD]) of data augmentation on the development set when increasingly adding synthesized data to the three learning systems, \ie functionals + SVMs (a), BoAWs + SVMs (b), and LLDs + GRU-RNNs (c). Experiments are repeated in 20 independent runs. scGANs: semi-supervised conditional GANs; cGANs: conditional GANs. net-60: default scGANs network structure with 60 nodes per hidden layer; average: averaged results over four different network structures of scGAN; ensemble: an ensemble of scGANs. }}
   \label{fig:res_da}
\end{figure*}

The dotted green curves in Fig.~\ref{fig:res_da} depict the obtained performance of the three learning systems as aforementioned when increasingly adding generated data to the original training set, by using the default scGANs architecture (\ie net-60; $N=60$ nodes per hidden layer). To mitigate the performance fluctuation caused by the random selection of generated data and the random initialization of neural networks, we repeated 20 independent runs for each data augmentation experiment. 

\begin{table*}[t!]
  \caption{Performance (UAR as well as corresponding Standard Deviation [SD]) comparison on both the development and test sets among the proposed scGAN-based data augmentation approaches, traditional data augmentation approaches, and the baseline systems in three learning systems, \ie functionals + SVMs, BoAWs + SVMs, and LLDs + GRU-RNNs. Experiments are repeated in 20 independent runs.  net-60: default scGANs network structure with 60 nodes per hidden layer; average: averaged results over four different network structures of scGANs; ensemble: an ensemble of scGANs.}
  \centering
  \begin{threeparttable}
  \begin{tabular}{llcc@{}lcc@{}lcc}
  \toprule
  approaches & & \multicolumn{2}{c}{functionals + SVMs} && \multicolumn{2}{c}{BoAWs + SVMs} && \multicolumn{2}{c}{LLDs + GRU-RNNs}  \\  
  \cline{3-4} \cline{6-7} \cline{9-10}
  UAR$_\text{SD}$ [\%] & & dev & test && dev & test && dev & test \\ 
  \midrule
  baseline (wo DA) & & $45.3_{\pm0.0}$ & $46.2_{\pm0.0}$ && $41.4_{\pm0.0}$ & $48.2_{\pm0.0}$ && $65.7_{\pm5.1}$ & $52.5_{\pm2.8}$ \\
  transformation~\cite{Amodei16-Deep} && $46.8_{\pm0.9}$ & $48.0_{\pm1.2}$ && $41.1_{\pm1.3}$ & $48.0_{\pm1.2}$ && $67.8_{\pm4.4}$ & $53.6_{\pm3.2}$ \\ 
  SMOTE\tnote{$^\star$}~\cite{Chawla02-SMOTE}	& & $45.1_{\pm0.5}$ & $47.0_{\pm0.8}$ && $41.3_{\pm0.5}$ & $47.9_{\pm1.1}$ && -- & -- \\ 
  \midrule
  {cGANs (net-60)} && $45.1_{\pm1.3}$ & $46.0_{\pm1.9}$ && $41.3_{\pm4.8}$ & $43.9_{\pm3.1}$ && $67.0_{\pm4.1}$ & $53.3_{\pm3.6}$  \\
  scGANs (net-60) & & $52.7_{\pm0.7}$ & $49.9_{\pm0.5}$ && $45.8_{\pm2.6}$ & $54.8_{\pm2.9}$ && $66.7_{\pm3.8}$ & $52.3_{\pm3.4}$ \\
  scGANs (average) & & $51.4_{\pm1.4}$ & $50.3_{\pm1.0}$ && $46.3_{\pm2.2}$ & $51.9_{\pm2.4}$ && $66.3_{\pm4.1}$ & $53.2_{\pm3.1}$ \\
  scGANs (ensemble) & & $53.8_{\pm2.4}$ & $\bf{51.5}_{\pm1.1}$ && $46.8_{\pm2.8}$ & $\bf{56.7}_{\pm3.4}$ && $67.4_{\pm4.0}$ & $\bf{54.4}_{\pm3.8}$ \\ 
  \bottomrule
  \end{tabular}
  \end{threeparttable}
  
    \begin{tablenotes}
    \scriptsize
    \item[] $\star$  SMOTE has not supported to synthesis sequences yet 
    \end{tablenotes}
    
  \label{tab:da}
\end{table*}

\begin{figure}[t!]
    \centering
    \includegraphics[width=2in]{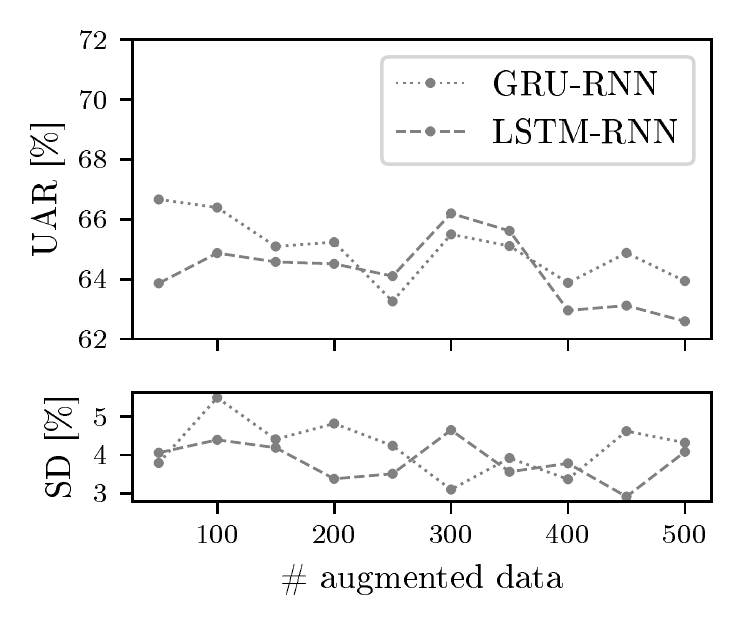}
    \vspace{-.2cm}
    \caption{{Performance comparison between LSTM-RNNs and GRU-RNNs for the sequence generation. }}
    \label{fig:res_lstm_gen}
    \vspace{-.5cm}
\end{figure}

In the case of the `functionals + SVMs' system (cf. Fig.~\ref{fig:res_da} (a)), it can be seen that the obtained UAR remarkably boosts from 45.3\,\% to 47.8\,\% when adding 50 synthesized samples per class, and dramatically to 52.7\,\% when adding 250 synthesized samples per class. Notable gain can also be observed for the `BoAWs + SVMs' system (\ie from 41.4\,\% to 45.9\,\% UAR). 
For the `LLDs + GRU-RNNs' system, a moderate improvement could be found (\ie from 65.7\,\% to 66.7\,\% UAR). This tells us that a scGAN-based data augmentation approach can indeed improve the performance of the systems when dealing with sparse data. {Besides, we compared this system with the LSTM-RNNs-based one with the same network architecture. The obtained results are shown in Fig.~\ref{fig:res_lstm_gen}. It can be seen that GRU-RNNs are competitive to the LSTM-RNNs, but with fewer parameters to be trained.}

When using the `functionals + SVMs' system, we see that its performance is continuously improving from the beginning but then remains almost stable when increasingly adding synthesized data. This may attribute to the fact that the model is prone to learn more from the synthesized data due to their higher weights. {Although increasing the capacity of a network may partially alleviate this problem, for the sake of better performance comparison, we retained the network architecture in all experimental scenarios.} For the `BoAWs + SVMs' and `LLDs + GRU-RNNs' systems, the maximum positive effect is achieved with a comparatively low number of only 50 synthesized samples per class, with a slight deterioration when adding more. A possible explanation is a mode collapse problem, where the generated data do not well reflect the whole picture of the original data distribution. 

Therefore, it is of importance to find the optimal balance between original and synthesized data. Such an observation, however, is not obvious and the ideal ratio might best be determined by experiments. 

We further notice that in the case of the `LLDs + GRU-RNNs' system, the data augmentation provides a limited performance enhancement. This underlines the toughness of generating sequential LLDs and requires further improvement in future efforts. 

{Moreover, we compared scGANs with cGANs, of which the performance is shown in Fig.~\ref{fig:res_da} with cyan curves. Obviously, scGANs are notably superior to cGANs in our case. This conclusion relates to the essential drawback of cGANs (cf.~Section~\ref{subsec:scGAN}). That is, although cGANs can simulate the overall data distribution, they are unable to guarantee the data distribution match for a particular snore sound category.}

\subsection{Results of the Ensemble of scGANs and Discussion} 
\noindent
To assess the effectiveness of an ensemble of scGANs, we conducted the experiments with four different network structures of scGANs, \ie N = 40, 60, 80, and 100. 
The black curves in Fig.~\ref{fig:res_da} illustrate the system performance through an ensemble of scGANs (\ie four scGANs).

Generally speaking, it can be seen that the ensemble of scGANs (\ie ensemble) outperforms the mono-scGAN (\ie net-60; dotted green curves) for data augmentation. The obtained UARs on the development set go up to 53.8\,\%, 46.2\,\%, and 67.4\,\%, respectively, in the cases of `functionals + SVMs', `BoAWs + SVMs', and `LLDs + GRU-RNNs'. 
We further averaged the performance of four mono-scGANs as outlined previously. Similar performance improvement of the ensemble of scGANs can be observed. Particularly, one can notice that more synthesized data are required to achieve the best performance when using an ensemble of scGANs. This implicitly indicates that the generated data are more diverse than the ones generated by a mono-scGAN, such that adding more generated data delivers better system performance. 

\begin{figure*}[t!]
  \centering
  \subfigure[{origin + synthesized data by scGAN (net-60)}]{
    \includegraphics[height=1.2in,trim={0.92cm 0.8cm 0.6cm 0.8cm},clip]{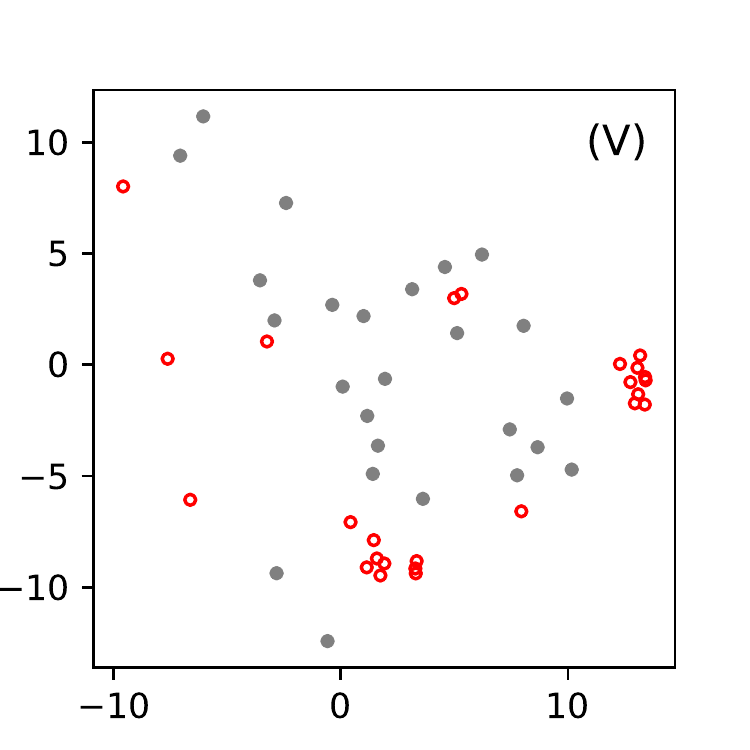} 
    \includegraphics[height=1.2in,trim={0.92cm 0.8cm 0.6cm 0.8cm},clip]{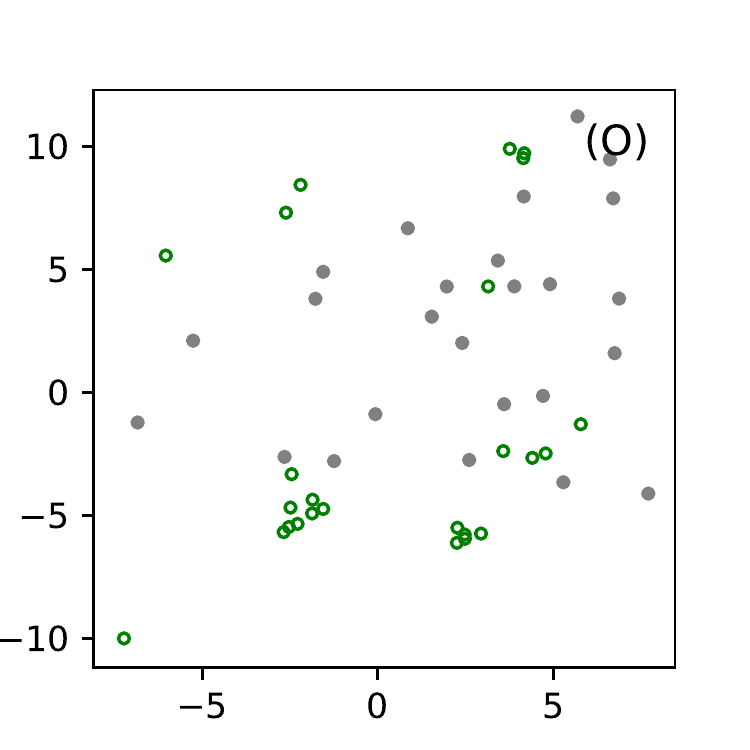}  
    \includegraphics[height=1.2in,trim={0.92cm 0.8cm 0.6cm 0.8cm},clip]{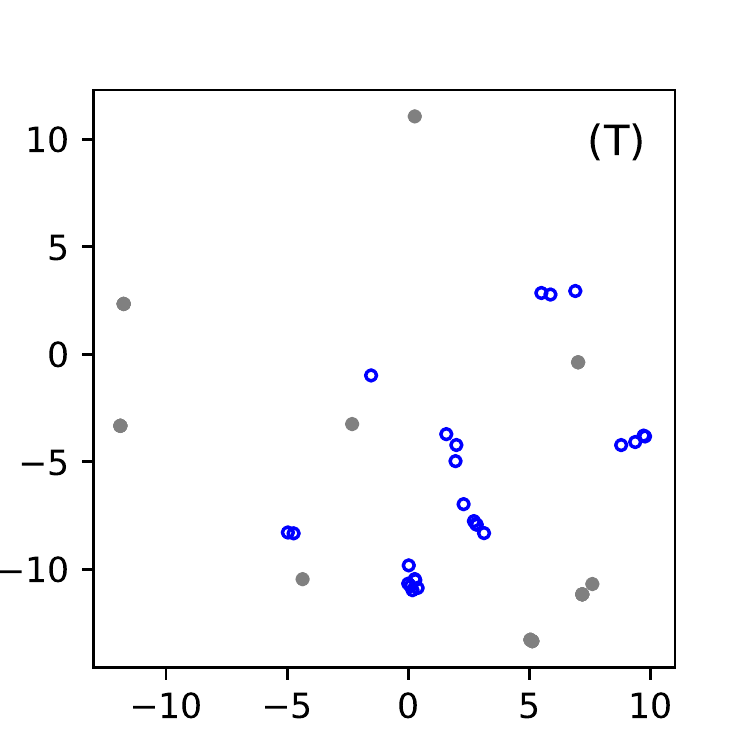}
    \includegraphics[height=1.2in,trim={0.92cm 0.8cm 0.6cm 0.8cm},clip]{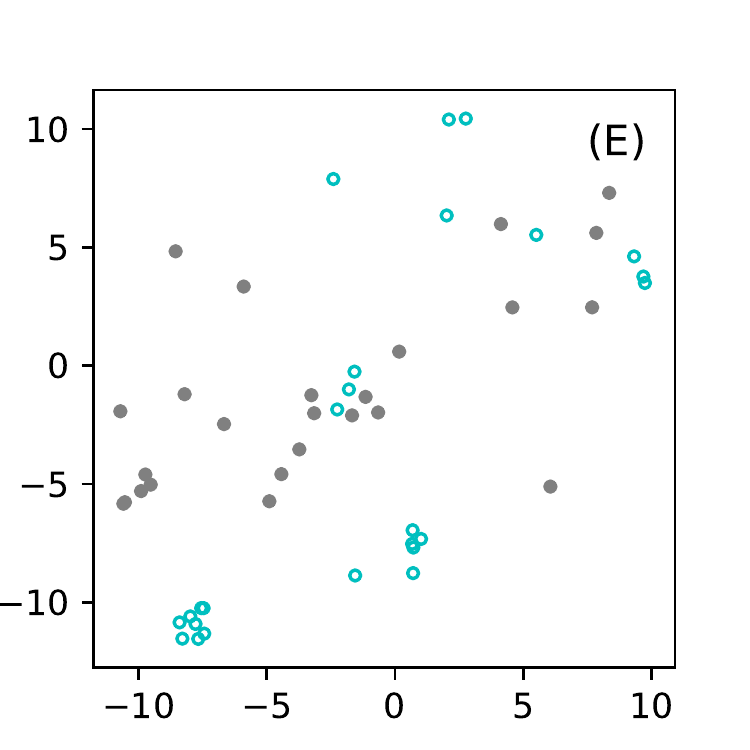}
    }
    \subfigure[{origin + synthesized data by scGAN (ensemble)}]{
    \includegraphics[height=1.2in,trim={0.92cm 0.8cm 0.6cm 0.8cm},clip]{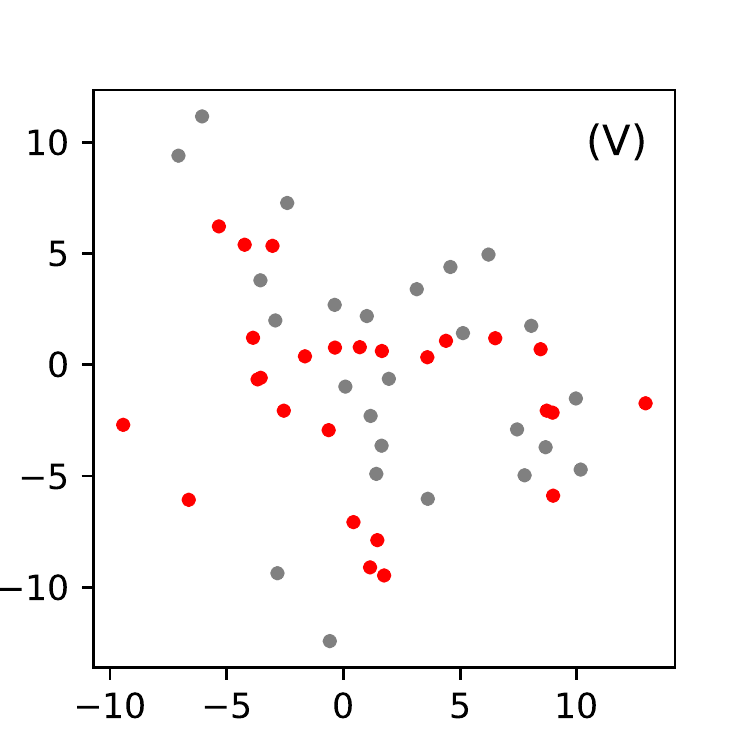}
    \includegraphics[height=1.2in,trim={0.92cm 0.8cm 0.6cm 0.8cm},clip]{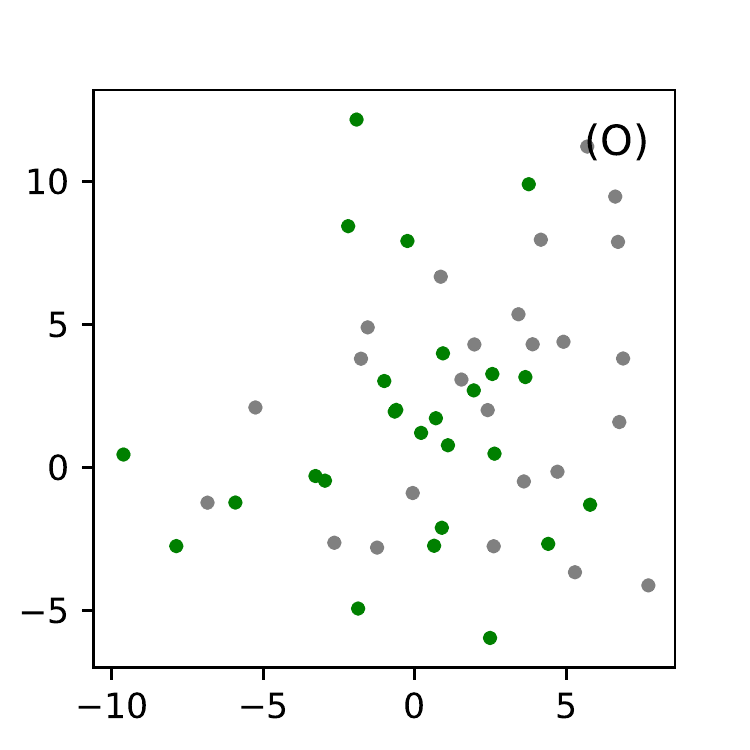}
    \includegraphics[height=1.2in,trim={0.92cm 0.8cm 0.6cm 0.8cm},clip]{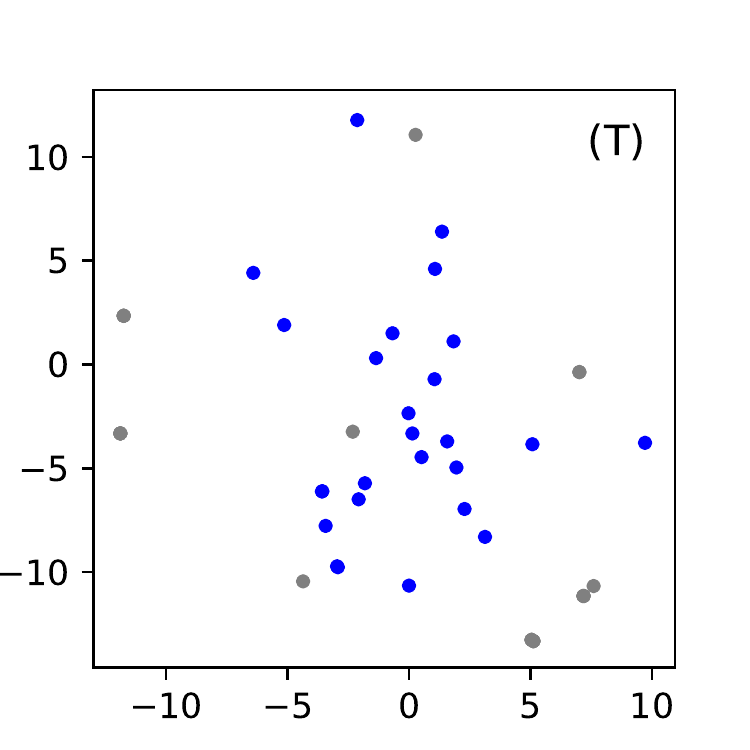}
    \includegraphics[height=1.2in,trim={0.92cm 0.8cm 0.6cm 0.8cm},clip]{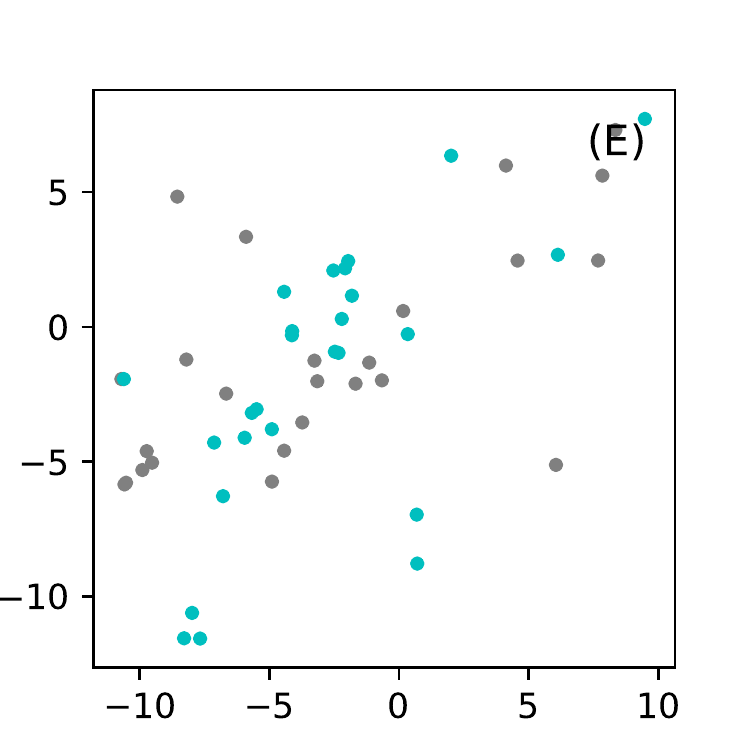}
    }
    \caption{{Visualization (t-SNE) of the mixed original and synthesized data (\ie functional-based features) in V, O, T, and E four categories. The synthesized data were generated through a mono-scGAN (net-60) (a) or an ensemble of scGANs (ensemble) (b). The grey points: original data; red, green, blue, and cyan circles/points: synthesized V, O, T, and E samples.}} 
  \label{fig:tSNE}
  \vspace{-.3cm}
\end{figure*}

{To intuitively demonstrate this conclusion, Fig.~\ref{fig:tSNE} illustrates the data distribution of the mixed original data and synthesized data (based on functionals), either simulated by a mono-scGAN (a) or ensemble of GANs (b). Generally speaking, compared with the mono-scGAN, the ensemble of scGANs is capable of generating more diverse data that better reflect the original data distribution in all V, O, T, and E cases. }
The data diversity, on the other hand, potentially results in less stable system performance, as shown in Fig.~\ref{fig:res_da} where the ensemble of scGANs generally shows higher standard deviations in 20 independent runs. 

Particularly, we investigated two classic data augmentation approaches, \ie audio transformation and SMOTE as briefly described in Section~\ref{sec:introduction}. For the transformation data augmentation, we degraded the original audio signals through diverse noises (\ie the CHiME noise~\footnote{obtained from \url{http://spandh.dcs.shef.ac.uk/chime_challenge/chime2016/data.html}} in rooms and the white noise) in different signal-to-noise ratios of $10\sim25$\,dB. In total,  data of ten times the size of the original set were generated. For the SMOTE approach, we synthesized the data belonging to the minority classes up to the number of the majority class (\ie 161 samples). 

In Table~\ref{tab:da} we compare the best UARs achieved on both the development and test sets through distinct approaches, \ie baseline systems without data augmentation, traditional data augmentation approaches, and the proposed scGAN-based approaches. 
Obviously, the results indicate that the scGAN-based data augmentation promotes the baseline systems without any data augmentation in the most scenarios. Furthermore, it is distinctly superior to the other two traditional data augmentation approaches. It can be seen that the SMOTE approach is merely competitive to the baseline, possibly because an upsampling operation has been applied to the original training set before the experiments (cf.~Section~\ref{subsec:setups}). Moreover, the system performance can be further improved when considering an ensemble of scGANs that is capable of dealing with the mode collapse problem of GANs.

\begin{table}[t!]
  \caption{{Performance comparison in terms of UAR between the proposed system with scGAN-based data augmentation and other state-of-the-art approaches on the MPSSC database. The experiments with sGANs and scGANs were conducted in 20 independent runs. }}
  \centering
  \begin{threeparttable}
  \begin{tabular}{lll}
  \toprule
  approaches (UAR$_\text{SD}$ [\%]) &  dev & test \\ 
  \midrule
  {\em related state-of-the-art approaches} \\ 
  \quad end-to-end~\cite{Tzirakis17-End} & 40.3 &  40.3 \\ 
  \quad fused end-to-end and BoAW + SVM~\cite{Schuller17-INTERSPEECH} & 45.1 & 46.0 \\   
  \quad dual source filter + SVM~\cite{Rao17-dual} &　49.6 & --- \\ 
  \quad fused GMM SV + SVM/RF, \\
  \quad \quad \quad \ \  Spec. + CNN~\cite{Nwe17-integrated} & 57.1 & 51.7 \\ 
  \quad fused FV/func. + (W)KPLS/KELM~\cite{Kaya17-Introducing} & --- & 64.2 \\ 
  \quad sGANs (functionals) &  $50.9_{\pm1.9}$ & $44.1_{\pm3.5}$ \\
  \quad sGANs (BoAWs) &  $49.0_{\pm2.7}$  & $51.1_{\pm3.9}$  \\
  \quad sGANs (LLDs) & $63.2_{\pm3.8}$ & $52.8_{\pm4.1}$ \\
  \midrule 
  \multicolumn{3}{l}{\em proposed snore-GANs (data augmentation by scGANs)} \\ 
  \quad functionals + SVMs & $53.8_{\pm2.4}$ & ${51.5}_{\pm1.1}$ \\
  \quad BoAWs + SVMs & $46.8_{\pm2.8}$ & ${56.7}_{\pm3.4}$ \\ 
  \quad LLDs + GRU-RNNs & $67.4_{\pm4.0}$ & $54.4_{\pm3.8}$ \\ 
  \bottomrule
  \end{tabular}
  \end{threeparttable}
  \label{tab:res_cmp}
  \vspace{-.3cm}
\end{table}

{
\subsection{Performance Comparison with Other State of the Art}
\label{subsec:res_cmp}
\noindent
To further compare the performance of our proposed data augmentation systems (snore-GANs) with other recently reported systems, we made a summary of the obtained UARs in Table~\ref{tab:res_cmp}. Generally speaking, it can be seen that our best-achieved results are competitive with, or even superior to, most of the other state-of-the-art systems. Particularly, we found that our systems can remarkably outperform the end-to-end system~\cite{Tzirakis17-End} (\ie 56.7\,\% vs 40.3\.\% UARs) that recently has been consistently regarded as one of the most attractive systems in the audio analysis~\cite{Tzirakis17-End}. This somewhat confirms the data sparsity challenge for the deep learning-based approaches that often prefer to a large amount of training data. 
Although some promising results were achieved in previous works, such as~\cite{Kaya17-Introducing}, i) the results on the development set were not provided; ii) the results were obtained by fusing several different systems, in contrast to our results delivered by merely one system. Furthermore, we also observe that the snore-GANs outperform the conventional sGANs in three different feature scenarios. This implies that the synthesized data indeed can help provide additional class-specific information for the classification models.}

\subsection{Discussion}
In future, we will keep collecting more snore sound data from different hospitals and patients to increase the data size and diversity, on which we will re-evaluate the proposed methods. 
Besides, more advanced or potential novel sequence generation systems (\eg variational recurrent autoencoders)~\cite{Press17-Language,Chung15-recurrent,Fabius14-Variational} will be further proposed and evaluated in our following work to improve the acoustic sequence generation models. Moreover, we intend to apply the approaches to other health care tasks (\eg cardiopathy and epilepsy) associated with other modalities, such as biological signals, images, and video recordings.

\section{Conclusion}
\label{sec:conclusion}
\noindent

To address the data scarcity problem for automatic snore sound classification (ASSC), we introduced a novel data augmentation approach based on semi-supervised conditional Generative Adversarial Networks (scGANs) in this article. 
By performing extensive experiments on the Munich-Passau snore sound corpus, we find that the scGANs-based data augmentation, especially its ensemble variation, is capable of generating new data which share a similar distribution with the original data, resulting in an increased quantity of training data without any human annotation efforts. By combining the synthesized and original data, the performance of ASSC systems was remarkably improved, indicating the effectiveness and robustness of the proposed approach for ASSC.

\ifCLASSOPTIONcaptionsoff
  \newpage
\fi

\bibliographystyle{IEEEtran}
\bibliography{refs}
\end{document}